\documentclass[letterpaper, 10 pt, conference]{ieeeconf}  

\IEEEoverridecommandlockouts                              

\usepackage{amssymb}  
\usepackage{bm}
\usepackage{booktabs}    
\usepackage{multirow}    
\usepackage{colortbl}    
\usepackage{xcolor}      
\usepackage{array}       
\usepackage{url}

\title{\LARGE \bf
AI-IO: An Aerodynamics-Inspired Real-Time Inertial Odometry for Quadrotors
}

\author{Jiahao Cui$^*$, Feng Yu$^*$, Linzuo Zhang, Yu Hu, and Danping Zou$^{\dag}$
\thanks{*These authors contributed equally to this work.}
\thanks{$^{\dag}$Corresponding author. All authors are with the Shanghai Key Laboratory of Navigation and Location-Based Services, Shanghai Jiao Tong University. This work was supported by National Key R\&D Program of China (2022YFB3903802) and National
Science Foundation of China (62073214).}%
\thanks{
{\tt\small Emails:\{csfalpha, yu-feng, zhanglinzuo, henryhuyu, dpzou\}@sjtu.edu.cn}}%
}

\usepackage{amsmath}
\usepackage{graphicx}
\usepackage{caption}

\begin{document}

\thispagestyle{empty}
\pagestyle{headings}

\makeatletter
\g@addto@macro\@maketitle{
  \captionsetup{type=figure}\setcounter{figure}{0}
  \def\mycolspace{1.2mm}
  \centering

}

\makeatother
\maketitle

\begin{abstract}
Inertial Odometry (IO) has gained attention in quadrotor applications due to its sole reliance on inertial measurement units (IMUs), attributed to its lightweight design, low cost, and robust performance across diverse environments. However, most existing learning-based inertial odometry systems for quadrotors either use only IMU data or include additional dynamics-related inputs such as thrust, but still lack a principled formulation of the underlying physical model to be learned. This lack of interpretability hampers the model’s ability to generalize and often limits its accuracy. In this work, we approach the inertial odometry learning problem from a different perspective. Inspired by the aerodynamics model and IMU measurement model, we identify the key physical quantity---rotor speed measurements required for inertial odometry and design a transformer-based inertial odometry. By incorporating rotor speed measurements, the proposed model improves velocity prediction accuracy by 36.9\%. Furthermore, the transformer architecture more effectively exploits temporal dependencies for denoising and aerodynamic modeling, yielding an additional 22.4\% accuracy gain over previous results. To support evaluation, we also provide a real-world quadrotor flight dataset capturing IMU measurements and rotor speed for high-speed motion. Finally, combined with an uncertainty-aware extended Kalman filter (EKF), our framework is validated across multiple datasets and real-time systems, demonstrating superior accuracy, generalization, and real-time performance. We share the code and data to promote further research (\url{https://github.com/SJTU-ViSYS-team/AI-IO}).
\end{abstract}

\section{INTRODUCTION}
 Low-cost IMUs are widely preferred for quadrotor state estimation owing to their compact size, low power consumption, and robustness to environmental degradation. However, achieving accurate inertial odometry with low-cost IMUs remains a critical challenge due to their high noise and bias instability.

Traditional inertial odometry, such as dead reckoning, suffers from long-term drift caused by random noise and slowly varying biases in IMUs. Consequently, significant efforts have been made to improve the accuracy of inertial odometry for specific domain applications. These methods typically employ explicit motion priors or learned motion patterns to impose constraints on inertial odometry estimation, such as zero velocity updates (ZUPT) for ground vehicles and pedestrians \cite{wang2015stance, li2023adaptive}. However, quadrotors are less constrained in their motion patterns, making it challenging to generalize such methods to quadrotor state estimation. Recent advances in inertial odometry for quadrotors have focused on learning-based methods. These works typically employ neural networks to predict position offsets or velocity estimates, yet they often fail to explicitly clarify the relationship between network inputs and outputs. For example, IMO \cite{cioffi2023learned} uses thrust instead of acceleration, and AirIO \cite{qiu2025airio} relies solely on IMU data and attitude. Based on aerodynamic model analysis \cite{sartori2018revisited, wang2017aerodynamic}, such input designs are inadequate, leading to incomplete observability in velocity estimation. The absence of a clear input-output mapping constrains prediction accuracy and generalization performance of these networks. Furthermore, most prior studies lack validation on real-world platforms.

\begin{figure}[t]
\centering
\includegraphics[width=0.5\textwidth]{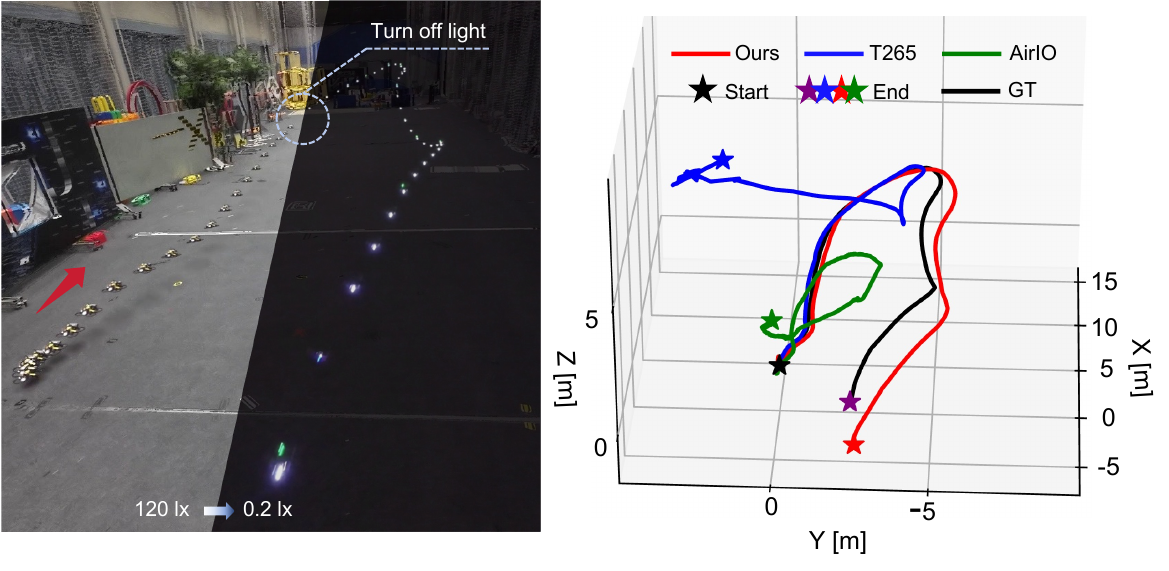}
	\caption{Comparison of our realtime AI-IO with Intel T265's visual-inertial odometry (VIO) and AirIO. Our method achieves comparable accuracy to VIO under normal lighting and superior performance under dark conditions, demonstrating stronger robustness to visual degradation, while also achieving higher accuracy than state-of-the-art AirIO.}
\label{fig:io-vio}
\end{figure}

In this work, we propose a novel perspective to explain and guide the design of quadrotor inertial odometry, along with a transformer-based velocity estimator. Our key insight is that augmenting IMU data with rotor speed enables the network to implicitly exploit the underlying quadrotor aerodynamics, thereby improving the accuracy of inertial odometry. Building on this, we introduce a lightweight transformer architecture—the first to be applied in quadrotor inertial odometry—which achieves improved velocity prediction accuracy while maintaining real-time inference efficiency. Extensive evaluations on both self-collected datasets and real-world flight experiments demonstrate that the proposed framework consistently outperforms existing approaches in terms of accuracy and generalization capability. In summary, our contributions are as follows:
\begin{itemize}
    \item We present an interpretable perspective that derives from the quadrotor aerodynamic model, offering a physics-based explanation of estimation errors and validating rotor speed as a critical measurement to enhance inertial odometry accuracy.
    \item Building on this perspective, we design the \textbf{Aerodynamics-Inspired Inertial Odometry (AI-IO)}, a lightweight transformer-based velocity estimator coupled with an uncertainty-aware EKF, enabling real-time, high-precision pose estimation.
    \item We establish a highly maneuverable quadrotor dataset using our self-built platform, including IMU data, rotor speed, and ground-truth poses and velocities, thereby filling a critical gap in datasets for agile quadrotor motions.
\end{itemize}

\section{RELATED WORK}
\subsection{Inertial Odometry}
Traditional model-based IO methods rely on integrating inertial measurements for state estimation, which is often unreliable because measurement errors and time-varying biases can cause significant accumulated drift in pose estimation. To mitigate the drift, IMUs are combined with exteroceptive sensors such as cameras \cite{qin2018vins} and LiDARs \cite{shan2020lio}. Although these algorithms achieve high accuracy, their dependence on external sensors makes them vulnerable to disturbances caused by agile motion or dynamic environments.

With the development of deep learning, data-driven methods have shown great potential for addressing noise and drift in low-cost IMU data, as well as for learning velocity or displacement directly from IMU measurements. Considering sensor errors, OriNet \cite{esfahani2019orinet} employs a LSTM network to calibrate gyroscope data, thereby improving the accuracy of inertial navigation. Similarly, Chen et al. \cite{chen2018improving} propose a deep learning-based approach to suppress multiple error sources in the sensor signals. Many other studies adopt a data-driven paradigm to estimate displacement or velocity directly from IMU data. For example, IONet \cite{chen2018ionet}, RoNIN \cite{herath2020ronin} and RIDI \cite{yan2018ridi} are end-to-end learning-based methods for pose estimation. Subsequent works integrate learned displacement or velocity into EKF frameworks \cite{liu2020tlio, sun2021idol, bajwa2024dive} or batched optimization frameworks \cite{buchanan2022learning}, aiming to build a more comprehensive state estimation pipeline. Recently, transformers have increasingly been adopted as the backbone of IO networks due to their powerful global attention mechanism. CTIN \cite{rao2022ctin} and iMoT \cite{nguyen2025imot} have demonstrated the superiority of transformer-based IO architectures. Most of the aforementioned IO studies focus on pedestrian motion, limiting their applicability to platforms with more demanding motion dynamics, such as quadrotors.

\subsection{Inertial Odometry for Quadrotors}
 Unlike pedestrian state estimation, quadrotors operate in six degrees of freedom in three-dimensional space, more complex motion patterns, and possess greater agility, making state estimation more challenging. The blade-element-momentum (BEM) theory \cite{bangura2016aerodynamics, gill2019computationally} combines blade-element theory and momentum theory to alleviate the difficulty of calculating induced velocity and to accurately capture the aerodynamic forces and torques acting on a single rotor under a wide range of operating conditions. Building on dynamic modeling of quadrotors, Sartori et al. \cite{sartori2018revisited} and Wang et al. \cite{wang2017aerodynamic} use the drag model to improve the quadrotor state estimation.

Although quadrotor aerodynamics and IMU observation model characterize the relationship between quadrotor body-frame velocity and IMU measurements, contemporary IO methods for quadrotors generally neglect aerodynamic drag models as a critical component. DIDO \cite{zhang2022dido} introduces a state estimation framework that integrates quadrotor dynamics and network observations in a two-stage EKF to jointly estimate IMU biases together with kinematic and dynamic states. IMO \cite{cioffi2023learned} employs collective thrust and gyroscope histories as inputs to a temporal convolutional network (TCN) to regress displacements, implicitly capturing quadrotor dynamics, but generalizing poorly to unseen trajectories due to its use of world-frame representation and the absence of acceleration inputs. Wang et al. \cite{wang2016velocity}, DIVE \cite{bajwa2024dive}, and AirIO \cite{qiu2025airio} rely solely on IMU data for prediction, disregarding rotor aerodynamic drag in the quadrotor dynamics model \cite{mahony2012multirotor}. As the current state-of-the-art, AirIO makes an important step forward by adopting body-frame representation. However, it still overlooks rotor speed as a key physical variable in velocity estimation, a limitation that our method explicitly addresses. In addition, all of these methods are evaluated on pre-collected datasets rather than deployed on resource-constrained quadrotor platforms for real-time tests.

\section{METHODOLOGY}
\subsection{Aerodynamics for Velocity Estimation}
\subsubsection{Notation}
The reference frame $\mathcal{W}$ is a fixed world coordinate system with its $z$-axis aligned with gravity. The quadrotor body frame and the IMU frame are denoted by $\mathcal{B}$ and 
$\mathcal{I}$, respectively, and all frames follow the right-hand convention. For any vector $\mathbf{x}$, the notation ${}^W\mathbf{x}$ denotes the representation of $\mathbf{x}$ in the $\mathcal{W}$ frame, with analogous notation applied to other frames. The position, velocity, and orientation of the body frame $\mathcal{B}$ with respect to the world frame $\mathcal{W}$ are denoted by ${}^W\mathbf{p}^W_B\in\mathbb{R}^3, {}^W\mathbf{v}^W_B\in\mathbb{R}^3,$ and $ \mathbf{R}^W_B\in\mathbb{R}^{3\times 3}$, respectively. For brevity, these quantities are hereafter denoted as $\mathbf{p}, \mathbf{v},$ and $ \mathbf{R}$ when the reference and target frames are clear from the context. Estimated and measured quantities are distinguished using the symbols $\hat{\mathbf{x}}$ and $\tilde{\mathbf{x}}$, respectively.
\subsubsection{IMU Model}
IMU measurements include the non-gravitational force in inertial space and the angular velocity of rigid frame given by:
\begin{equation}
\begin{aligned}
^{I}\tilde{\boldsymbol{\omega}} &= {}^{I}\boldsymbol{\omega} 
+ {}^{I}\mathbf{b}_{g} + \mathbf{n}_{g}\\
^{I}\tilde{\mathbf{a}} &= {}^{I}\mathbf{a} 
+ {}^{I}\mathbf{b}_{a} 
- ({\mathbf{R}^{W}_{I}})^{\top} \, {}^{W}\mathbf{g} 
+ \mathbf{n}_{a}
\end{aligned}
\label{imu-model}
\end{equation}
where $^{I}\tilde{\boldsymbol{\omega}}$ and $^{I}\tilde{\mathbf{a}}$ are the measurements of on-board gyroscope and accelerometer, ${}^{I}\boldsymbol{\omega}$ and ${}^{I}\mathbf{a} $ are the ground-truth angular velocity and acceleration, $\mathbf{n}_{g}$ and $\mathbf{n}_{a}$ are Guassian noise, and ${}^{W} \mathbf{g}=[0, 0, -9.81]$ ($m/s^2$) is the gravity vector, ${}^{I}\mathbf{b}_{g}$ and ${}^{I}\mathbf{b}_{a}$ are biases modeled as random walk processes with noise $\mathbf{n}_{\mathbf{b}_g}$ and $\mathbf{n}_{\mathbf{b}_a}$.

\subsubsection{Quadrotor Aerodynamics}
A quadrotor has four propellers fixed in a rigid body symmetrically, with one pair of propellers rotating counter-clock-wise and the other pair clock-wise.
To reveal the relationship between accelerometer measurements and quadrotor velocity, we model the observation equation \eqref{imu-model} of IMU sensor combined with aerodynamics of quadrotor following \cite{sartori2018revisited,wang2017aerodynamic} as follows:
\begin{equation}
\left\{
\begin{aligned}
\tilde{a}_x &= \underbrace{- \lambda_x \omega_m^2 v_x}_{induced \: drag} \underbrace{-k_1 v_x - k_2 v_x^2}_{air \: drag} + \underbrace{b_{a_x} + n_{a_x}}_{noise}, \\
\tilde{a}_y &= - \lambda_y \omega_m^2 v_y -k_3 v_y - k_4 v_y^2 + b_{a_y} + n_{a_y}, \\
\tilde{a}_z &= \underbrace{4\alpha{\omega_m^2}}_{total \: thrust} - k_5 v_z - k_6 v_z^2 + b_{a_z} + n_{a_z}
\end{aligned}
\right.
\label{drone_dynamics}
\end{equation}
where $w_m^2=\frac{\sum_{i=1}^4 \omega_i^2}{4}$, $\omega_i$ denotes the rotor speed of the $i$-th rotor. $k_i$, i=1,2,...,6, $\lambda_j$, j=\{x,y\}, and $\alpha$ are approximately constant parameters depend on material, structure of propellers and airframe, and air density. Note that aerodynamic forces depend on the relative airflow velocity rather than the body-frame velocity. However, since we consider windless conditions in this work, the two are equivalent and the above formulation is valid. Further considering the effect of the non-inertial frame, the Coriolis acceleration term
$\boldsymbol{a}_c=-2\boldsymbol{\omega \times \boldsymbol{v}}$
is added to \eqref{drone_dynamics} to compensate for quadrotor's rotation, which explains why angular velocity is required as one of the observations for network inputs.

\subsubsection{Rotor Speed}
According to \eqref{drone_dynamics}, if the environment and quadrotor body properties remain constant over a short time horizon and noise is neglected, IMU measurements and body velocity are mutually derivable under our modeled aerodynamics when other observations (e.g., rotor speeds) are available. Therefore, assuming that the network can model the aerodynamics from temporal observations (i.e., coefficients such as $k_{i=1,2,...,6}, \lambda_{j=x,y}$ and $\alpha$ in \eqref{drone_dynamics}), beyond the acceleration and angular velocity measured by IMU, we argue that introducing rotor speeds as inputs to ensure observability of the estimator is critical to enhancing velocity estimation accuracy—a claim validated by subsequent experiments. This property forms the basis of our method.

\begin{figure}
    \centering
    \includegraphics[width=\linewidth]{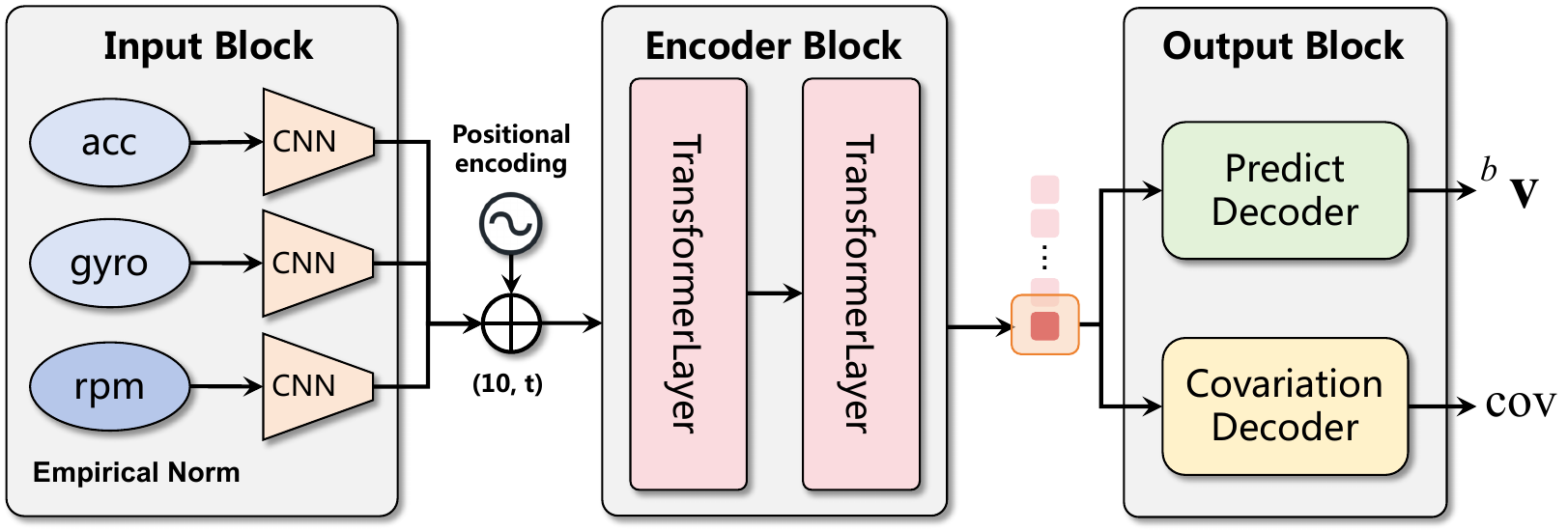}
    \caption{Proposed transformer-based architecture: Verified measurements are normalized, encoded, and fed into a transformer to capture spatiotemporal dependencies. Finally, two fully connected decoders predict velocity and uncertainty.}
    \label{fig:network}
\end{figure}

\subsection{Network Design \& Training}
Although the proposed equation \eqref{drone_dynamics} indicates a single-frame correspondence between IMU measurements and quadrotor speed, prior studies have employed historical multi-frame data buffer for estimation \cite{bajwa2024dive,zhang2022dido,cioffi2023learned,qiu2025airio}. We attribute this to the use of historical data buffer, which aids in better estimating IMU noise and better aerodynamics modeling. In recent years, transformer networks have demonstrated superior denoising capabilities across various modal data \cite{yang2024denoising,wang2022uformer,chen2020dual}, so we integrate a transformer network with our input and output blocks to process historical data buffer, as shown in Fig. \ref{fig:network}.

Directly feeding raw noisy data into the encoder may fail to capture the complex relationships between individual feature dimensions. Therefore, we employ CNNs to process raw sensor data sequences. We posit that applying multiple temporal convolutions acts as a low-pass filter, removing specific high-frequency noise \cite{brossard2020denoising}. Furthermore, the local inductive bias inherent in CNNs aligns with the local correlation characteristics of IMU data, enabling effective extraction of local features \cite{wang2023theoretical}. Note that the rotor speed values are often much larger than other measurements; thus, we normalize the mean and variance of rotor speeds online using empirical values to align with the standard normal distribution and enhance training stability.

After fusion with position encoding, the processed features is fed into a transformer for temporal feature extraction and aerodynamics modeling. Unlike the original transformer, we eliminate the attention mechanism in its decoder and directly employ two fully connected layers to process the encoder’s temporal features of the last moment. These fully connected layers independently generate the predicted velocity and uncertainty of the last frame, enabling lightweight and real-time inference. 

To supervise the velocity prediction and uncertainty estimation of the last frame, we adopt two loss functions \cite{liu2020tlio, russell2021multivariate}:
\begin{equation}
    \begin{aligned}
        \mathcal{L}_V &= \left\{
        \begin{array}{ll}
          \frac{1}{2}(^{B}\mathbf{v}_i - ^{B}\mathbf{\hat{v}}_i)^2, & \text{if $\lvert ^{B}\mathbf{v}_i - ^{B}\mathbf{\hat{v}}_i \rvert < \delta$,} \\
          \delta \cdot (\lvert ^{B}\mathbf{v}_i - ^{B}\mathbf{\hat{v}}_i\rvert - \frac{1}{2}\delta), & \text{otherwise}
        \end{array}
        \right. \\
            \mathcal{L}_C &= \left( ^{B} \mathbf{v}_i - ^{B}\mathbf{\hat{v}}_i \right)^\top \hat{\Sigma}_i^{-1} \left( ^{B} \mathbf{v}_i - ^{B}{\mathbf{\hat{v}}}_i \right) + \ln \left( \det \hat{\Sigma}_i \right)
    \end{aligned}
\end{equation}
where the $\mathcal{L}_V$ is the Huber loss between the ground-truth velocity $^{B}\mathbf{v}_i$ and the predicted velocity $^{B}{\mathbf{\hat{v}}_i}$ for the $i$-th data inputs, the $\mathcal{L}_C$ is the negative-log-likelihood (NLL) loss used for uncertainty supervision, $\hat{\Sigma}_i=diag(\hat{\sigma}_i)$ is the predicted uncertainty in the form of covariance matrix with only diagonal values $\hat{\sigma}_i$ output by the decoder. Once the velocity prediction loss has converged, training proceeds with the NLL loss term as suggested by \cite{bajwa2024dive}.

\subsection{Dataset Preparation}
\begin{figure}
    \centering
    \includegraphics[width=\linewidth]{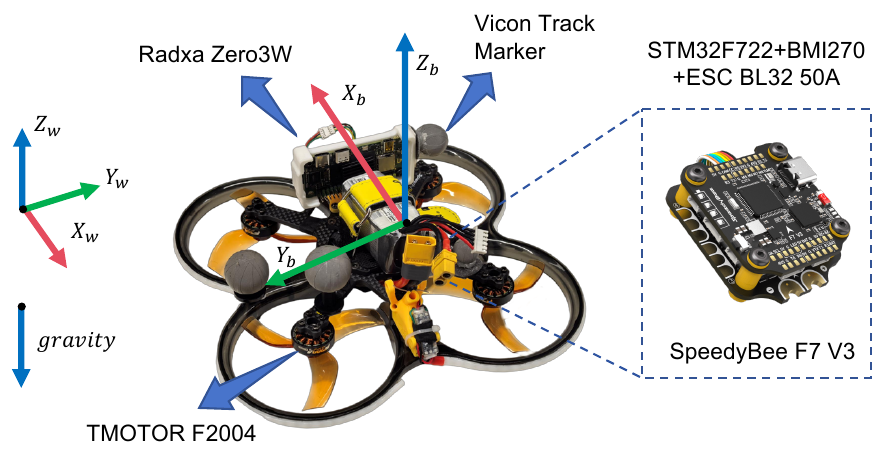}
    \caption{Hardware setup which assembles a low-cost BMI270 IMU for acceleration and angular velocity and ESCs using bidirectional Dshot for rotor speed.}
    \label{fig:dataset-platform}
\end{figure}

Among the currently available quadrotor datasets, only the Blackbird \cite{Amado2018blackbird}, VID \cite{Zhang2021vid}, and DIDO \cite{zhang2022dido} datasets include rotor speeds. However, the flight trajectories in the Blackbird are overly repetitive, limiting network generalization. The VID and DIDO primarily contain non-aggressive flight data with relatively low speeds. To meet the requirements of this study for high-aggressiveness flight data with rotor speed, we equip a quadrotor as a data acquisition platform (shown in Fig. \ref{fig:dataset-platform}) and collect a large amount of high-speed flight data. The quadrotor is built on a SpeedyBee\footnote{https://www.speedybee.com/} Bee35 chassis with an F7 V3 Stack flight controller. IMU data is obtained from the BMI270\footnote{https://www.mouser.sg/new/bosch/bosch-sensortec-bmi270/} sensor, rotor speed is recorded via telemetry feedback from the electronic speed controller, and the ground truth is provided by a VICON\footnote{https://www.vicon.com/} system.

To capture flight data with a diverse range of speeds, we manually control the quadrotor to fly at three levels: low, medium, and high, with maximum speeds of approximately 5m/s, 8m/s, and 14m/s, respectively. Four sequences are recorded for each speed, resulting in a total of 12 sequences, covering a flight distance of 6,704m and a duration of 1,960s. In addition, we collect low-speed data (below 3 m/s) from autonomous flights. These include planar circular and figure-eight patterns, non-planar circular and figure-eight patterns, and random flight trajectories. Two sequences are recorded for each type, yielding a total of 10 sequences, covering a flight distance of 563m and a duration of 676s.

\subsection{Extended Kalman Filter}

\subsubsection{System State}
Our EKF builds upon the framework established in \cite{qiu2025airio}. The state at time $i$ is defined as $\mathbf{X}_i = \{\hat{\mathbf{R}}_i, \hat{\mathbf{v}}_i, \hat{\mathbf{p}}_i, \hat{\mathbf{b}}_{a_i}, \hat{\mathbf{b}}_{g_i}\}$. Following the error-state formulation in \cite{Mourikis2007multistate}, the corresponding error-state vector is defined as $\delta \mathbf{X}_i = \{\delta\bm{\theta}_i, \delta\mathbf{v}_i, \delta\mathbf{p}_i, \delta\mathbf{b}_{a_i}, \delta\mathbf{b}_{g_i}\}$. For Euclidean vector variables $\mathbf{x}\in\mathbb{R}^3$, the error vector is given by $\delta\mathbf{x} = \mathbf{x} - \hat{\mathbf{x}}$. For rotational components, the minimal error $\delta\bm{\theta}\in\mathbb{R}^3$ is obtained via the logarithmic map $\delta\bm{\theta} = \text{Log}(\mathbf{R}\cdot\hat{\mathbf{R}}^{-1})$ whose inverse is given by the exponential map $\text{Exp}(\delta\bm{\theta})$. 

\subsubsection{State Propagation}
We propagate the state using high-frequency IMU measurements via:
\begin{equation}
\left\{
\begin{aligned}
\hat{\mathbf{R}}_{i+1} &= \hat{\mathbf{R}}_i \cdot \text{Exp}(\tilde{\bm\omega}_i - \hat{\mathbf{b}}_{g_i})\cdot\Delta t\\
\hat{\mathbf{v}}_{i+1} &= \hat{\mathbf{v}}_i + {}^W\mathbf{g} \cdot \Delta t +  \hat{\mathbf{R}}_i \cdot (\tilde{\mathbf{a}}_i - \hat{\mathbf{b}}_{a_i}) \cdot \Delta t\\
\hat{\mathbf{p}}_{i+1} &= \hat{\mathbf{p}}_i + \hat{\mathbf{v}}_i \cdot \Delta t +  \frac{1}{2} \cdot [{}^W\mathbf{g} + \hat{\mathbf{R}}_i (\tilde{\mathbf{a}}_i - \hat{\mathbf{b}}_{a_i})] \cdot \Delta t^2\\
\hat{\mathbf{b}}_{a_{i+1}} &= \hat{\mathbf{b}}_{a_i},
\hat{\mathbf{b}}_{g_{i+1}} = \hat{\mathbf{b}}_{g_i}
\end{aligned}
\right.
\label{INS}
\end{equation}
where $\Delta t$ denotes the integration interval.
The linearized error-state and covariance propagation are modeled as:
\begin{equation}
\begin{aligned}
    \delta \mathbf{X}_{i+1} = \mathbf{A}_i \cdot \delta \mathbf{X}_i + \mathbf{B}_i \cdot \mathbf{n}_i
\end{aligned}
\label{linear_prop_model}
\end{equation}
\begin{equation}
    \begin{aligned}
        \mathbf{P}_{i+1} = \mathbf{A}_i \cdot \mathbf{P}_i \cdot \mathbf{A}_i^\top + \mathbf{B}_i \cdot \mathbf{W}_i \cdot \mathbf{B}^\top_i,
    \end{aligned}
    \label{linear_prop_cov_model}
\end{equation}
where $\mathbf{n}_i = [\mathbf{n}_{a_i}, \mathbf{n}_{g_i}, \mathbf{n}_{\mathbf{b}_{a_i}}, \mathbf{n}_{\mathbf{b}_{g_i}}]^\top$ represents sensor noise and bias noise, $\mathbf{W}_i$ is the propagation noise matrix.

\subsubsection{Measurement Update}
The observation model based on velocity in $\mathcal{B}$ is:
\begin{equation}
    \begin{aligned}
        {}^B\mathbf{v}^W_B =h(\mathbf{X}) = (\mathbf{R}^W_b)^\top \cdot {}^W\mathbf{v}^W_B
    \end{aligned}
    \label{observation_model}
\end{equation}
where $\mathbf{n}_\mathbf{v}$ denotes zero-mean Gaussian noise on the predictions of the network, with its covariance also provided by the network. Linearizing \eqref{observation_model}, the Jacobian matrix $\mathbf{H}$ is straightforward to compute, with all entries zero except
\begin{equation}
    \begin{aligned}
        \mathbf{H}_{\bm\theta} &= \frac{\partial h(\mathbf{X})}{\partial \delta \bm\theta} = \hat{\mathbf{R}}^\top \lfloor \hat{\mathbf{v}} \rfloor_\times, \\
        \mathbf{H}_{\mathbf{v}} &= \frac{\partial h(\mathbf{X})}{\partial \delta \mathbf{v}} = \hat{\mathbf{R}}^\top
    \end{aligned}
\end{equation}
where $\lfloor \bm\theta \rfloor_\times$ denotes the skew-symmetric matrix of vector $\bm\theta$.

Finally, $\mathbf{H}$ are used to compute the Kalman gain to update the state $\mathbf{X}$ and the covariance $\mathbf{P}$ as follows:
\begin{equation}
\left\{
    \begin{aligned}
        \mathbf{K} &= \mathbf{P} \cdot \mathbf{H}^\top \cdot (\mathbf{H} \cdot \mathbf{P} \cdot \mathbf{H}^\top + \bm\Sigma)^{-1}\\
        \mathbf{X} &\leftarrow \mathbf{X} + \mathbf{K} \cdot ({}^B\tilde{\mathbf{v}} - \hat{\mathbf{R}}^\top \cdot \hat{\mathbf{v}})\\
        \mathbf{P} &\leftarrow (\mathbf{I} -\mathbf{K} \cdot \mathbf{H}) \cdot \mathbf{P} \cdot (\mathbf{I} -\mathbf{K} \cdot \mathbf{H})^{-1} + \mathbf{K} \cdot \bm\Sigma \cdot \mathbf{K}^\top
    \end{aligned}
\right.
\end{equation}
where $\bm\Sigma$ is the measurement covariance provided by the network.

During filter initialization, the insufficient data length prevents network inference, restricting the system to inertial propagation. Once sufficient data accumulates, the network performs inference according to the filter update frequency to provide body-frame velocity estimates.

\section{EXPERIMENTAL RESULTS}
We evaluate our method on two quadrotor datasets: the relatively low-speed DIDO \cite{zhang2022dido} dataset and our own customized aggressive high-speed dataset. Both datasets provide rotor speed measurements, enabling us to verify the effectiveness of the proposed algorithm under different maneuvering conditions. We compare our system with the following baselines:
\begin{itemize}
    \item IMU preintegration \cite{forster2015imu}. A model-based method that pre-integrates the IMU measurements on the SO(3) manifold and simultaneously efficiently corrects the IMU bias estimate online.
    \item IMO \cite{cioffi2023learned}. A learning-based inertial odometry approach that embeds control signals specifically designed for drone racing, with high positioning accuracy on a fixed track trajectory.
    \item AirIO \cite{qiu2025airio}. The current state-of-the-art learning-based inertial odometry algorithm for quadrotors, which validates the effectiveness of predicting body velocity directly from IMU data.
\end{itemize}
Among these methods, IMU preintegration represents the model-based approach, while IMO and AirIO are representative learning-based methods for quadrotors.

\subsection{Metrics Definitions}
To assess the performance of our system, we adopt two widely used metrics to quantify the localization performance \cite{herath2020ronin, liu2020tlio, qiu2025airio}: Absolute Translation Error (\textbf{ATE}, m) computes the Root Mean Squared Error (RMSE) between the estimated and ground-truth velocities over all
time points, and Relative Translation Error (\textbf{RTE}, m) computes the RMSE of the relative velocities over predefined time intervals. In our experiments, we adopt a 5-second interval. Similar to position, we use Absolute Velocity Error (\textbf{AVE}, m/s) and Relative Velocity Error (\textbf{RVE}, m/s) to evaluate the accuracy of the velocity estimation.

\subsection{Ablation Studies}
\begin{figure}
    \centering
    \includegraphics[width=\linewidth]{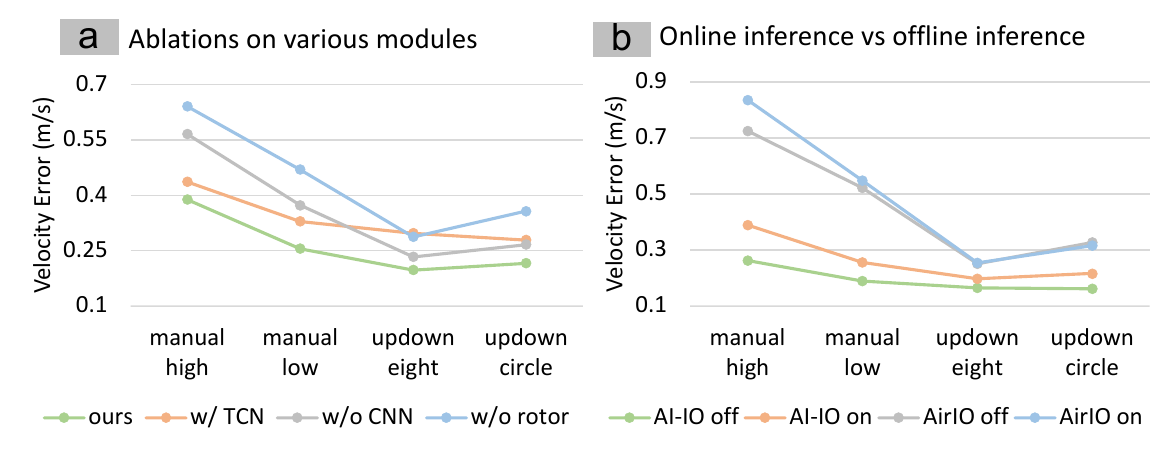}
    \caption{(a) Ablation results for each module on the test dataset. (b) Comparison of online and offline inference results between AI-IO and AirIO with a window length set to 1 second.}
    \label{fig:ablation}
\end{figure}
\subsubsection{Ablations on each module}
We conduct a series of ablation experiments to analyze the effects of different components on the network, with a focus on assessing the influence of key observed values from rotor speed measurements. All experiments are performed using our self-collected dataset, with results presented in the Fig. \ref{fig:ablation} (a):
\begin{itemize}
    \item \textbf{Model Backbone}: Compared with TCN, using transformer as the network backbone reduces the average body-frame velocity prediction error on the test set by 22.4\%. We attribute this reduction to transformer’s enhanced ability to model temporal dependencies, such as estimating biases and capturing aerodynamic parameters from historical observations.
    \item \textbf{Feature Extractor}: Placing CNN before the transformer backbone enables denoising and feature extraction, leading to a 23.9\% reduction in the network’s average velocity prediction error. This effect is particularly obvious on high-speed trajectories with higher noise.
    \item \textbf{Rotor Speed}: As with the aerodynamic modeling discussed in \eqref{drone_dynamics}, introducing rotor speed measurements ensures the system’s observability. Therefore, we compare velocity predictions without rotor speeds and observe a 36.9\% reduction in velocity prediction error. This experimental result provides a better explanation for neural network learning.
\end{itemize}

\subsubsection{Online inference vs offline inference}
Although offline inference is not practically applicable in real-world scenarios, due to AirIO’s excellent offline inference performance \cite{qiu2025airio}, we conduct a comprehensive comparison and analysis of velocity prediction errors between offline and online inference. 

The key distinction between online and offline inference lies in how the sliding window is updated: online inference shifts the window by one time step, while offline inference shifts the entire window. Additionally, the supervised signals for both methods are derived from either the velocity at the window’s end (online) or the velocities at multiple time steps within the window (offline). To validate this, we set the window length to 1 second. Fig. \ref{fig:ablation} (b) shows that offline inference achieves lower average velocity prediction error on its trajectory. This improvement stems from the denser supervision signals within the full window and its use of global window updates, with the effect being particularly pronounced on AI-IO. In order to fully compare the accuracy of real-time processing, the subsequent comparisons on datasets all use a 1-second window for online inference.
\subsection{Comparison on Datasets}

\subsubsection{DIDO Dataset}
\begin{table*}[ht]
\caption{Comparison on DIDO datasets. The best and second best results are bold and underlined respectively.}
    \centering
    \setlength{\tabcolsep}{4.5pt}
    \renewcommand{\arraystretch}{1.2}
        \begin{tabular}{l|cccc|cccc|cccc|cccc}
        \toprule
        \multirow{2}{*}{\textbf{Sequence}} & 
        \multicolumn{4}{c}{\textbf{AirIO Net}$^1$} & 
        \multicolumn{4}{c|}{\textbf{AI-IO Net}$^1$} & 
        \multicolumn{4}{c}{\textbf{AirIO EKF}} & 
        \multicolumn{4}{c}{\textbf{AI-IO EKF}} \\
        & AVE & RVE & ATE & RTE & AVE & RVE & ATE & RTE & AVE & RVE & ATE & RTE & AVE & RVE & ATE & RTE \\
        \midrule
        circle     & 0.588 & 0.647 & 4.827 & 1.146 & \textbf{0.327} & \textbf{0.364} & \textbf{2.588} & \textbf{0.724} & 1.129 & 1.229 & 6.600 & 2.150 & \underline{0.385} & \underline{0.432} & \underline{3.333} & \underline{0.915} \\
        updown circle   & 0.563 & 0.742 & \underline{4.999} & \underline{1.153} & \textbf{0.321} & \textbf{0.403} & \textbf{3.608} & \textbf{0.836} & 1.960 & 1.126 & 6.085 & 1.806 & \underline{0.355} & \underline{0.427} & 5.319 & 1.158 \\
        eight   & 0.506 & 0.733 & \underline{2.610} & 1.252 & \textbf{0.321} & \textbf{0.392} & \textbf{1.448} & \textbf{0.684} & 0.870 & 1.239 & 5.417 & 2.442 & \underline{0.396} & \underline{0.532} & 2.833 & \underline{1.105} \\
        updown eight   & 0.639 & 0.940 & 3.535 & 1.517 & \textbf{0.407} & \textbf{0.474} & \textbf{1.750} & \textbf{0.753} & 1.057 & 1.443 & 6.194 & 2.766 & \underline{0.431} & \underline{0.593} & \underline{2.419} & \underline{1.167} \\
        random   & 0.522 & 0.772 & \textbf{3.553} & 1.294 & \textbf{0.278} & \underline{0.363} & \underline{3.788} & \textbf{0.721} & 0.895 & 1.366 & 5.564 & 2.408 & \underline{0.287} & \textbf{0.342} & 5.474 & \underline{0.975} \\
        \midrule
        \textbf{Average}     & 0.564 & 0.767 & 3.905 & 1.272 & \textbf{0.331} & \textbf{0.399} & \textbf{2.636} & \textbf{0.744} & 1.182 & 1.281 & 5.972 & 2.314 & \underline{0.371} & \underline{0.465} & \underline{3.876} & \underline{0.903} \\
        \bottomrule
        \multicolumn{17}{l}{1 leverage ground truth orientation to transform the net output to world frame for inference.} \\
    \end{tabular}
    \label{DIDO comparison}
\end{table*}
We evaluate our method AI-IO and the state-of-the-art method AirIO on the DIDO dataset. We use all random flight data starting with "v" for training, and perform validation and testing on other periodic flight data and random flight data. 

Table \ref{DIDO comparison} shows the evaluation results on 5 types of sequences. The results of each type of sequence are the average of multiple sequences. AirIO Net and AI-IO Net use the real attitude information to convert the network output from body frame to world frame, and then directly integrate to obtain the position, while AirIO EKF and AI-IO EKF use attitude estimates instead of the true attitude. Because observation updates rely on body-frame velocity, errors in attitude estimates, particularly those introduced by low-cost IMUs such as the BMI270 we employed, can propagate to velocity updates and subsequently degrade position accuracy. Therefore, AI-IO EKF decreases 12.1\% and 47.0\% in AVE and ATE respectively compared to AI-IO Net. Since AirIO uses attitude as inputs, it introduces larger errors when the attitude estimation is inaccurate, resulting in a 109.6\% and 52.9\% decrease in AVE and ATE, respectively. Overall, AI-IO Net improves AVE and ATE by 41.3\% and 32.5\% respectively compared to AirIO Net, and AI-IO EKF improves AVE and ATE by 68.6\% and 35.1\% respectively compared to AirIO EKF.

\subsubsection{Our Dataset}

In order to fully verify the generalization of AI-IO across different speed distributions and different flight trajectories, we use our own collected data for training and evaluation. All manual flight data sequences are divided into training, validation, and testing data sets according to the ratio of 7:1.5:1.5, and all non-manual flight data are divided into the testing set. To avoid the influence of the filter parameter settings, the results of the unfiltered inference of AirIO and AI-IO are compared here. IMO's network output is a displacement within a window, which makes it difficult to directly infer velocity and position, so its original filter fusion is still used.

\begin{figure}
    \centering
    \includegraphics[width=\linewidth]{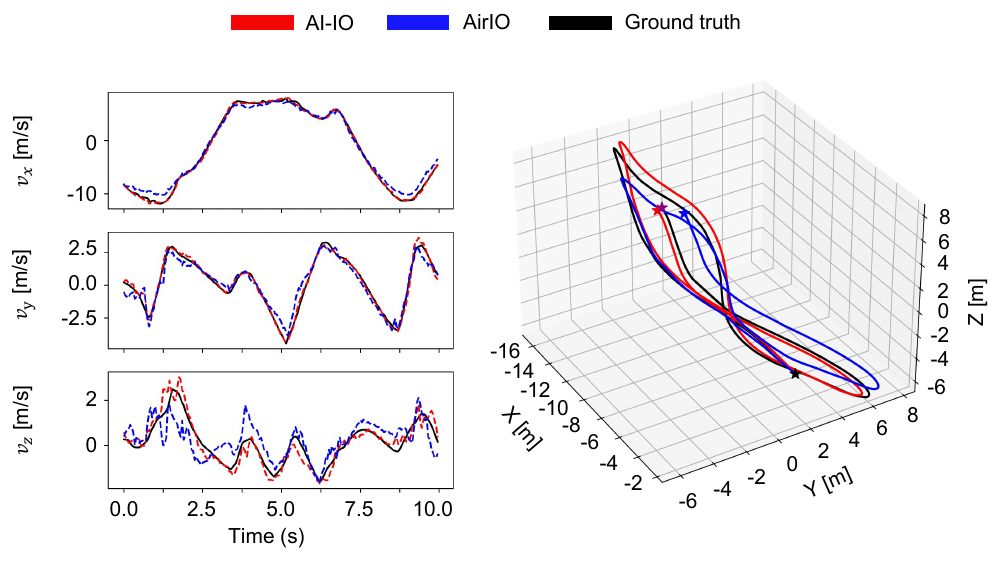}
    \caption{Comparison of estimated velocity and trajectory in a manual high sequence, with the maximum speed exceeding 10m/s. The ATEs of AI-IO and AirIO are 0.860 and 1.762 respectively, and AI-IO outperforms AirIO by 51.2\% in ATE.}
    \label{fig:high-spd-traj}
\end{figure}

\begin{figure}
    \centering
    \includegraphics[width=\linewidth]{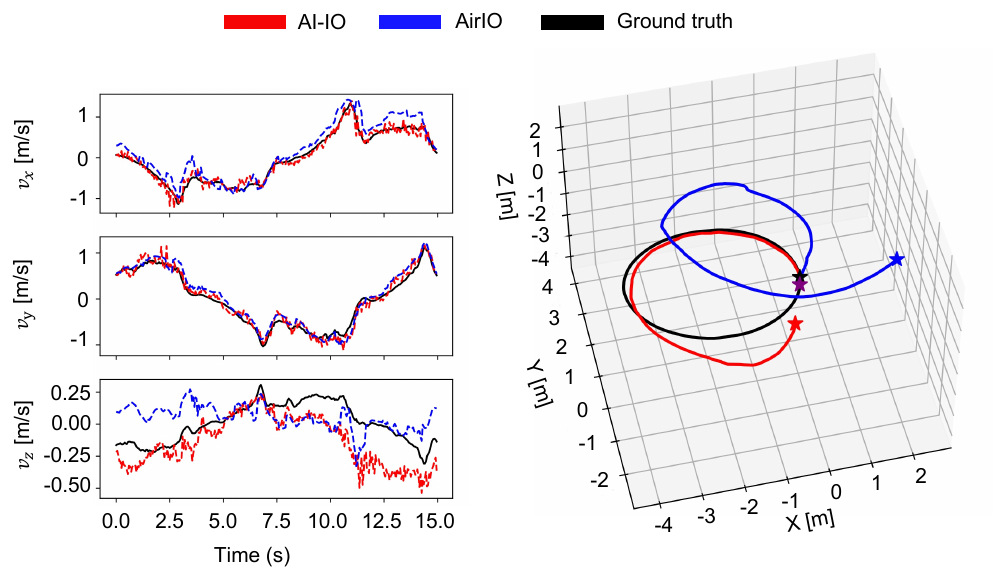}
    \caption{Comparison of estimated velocity and trajectory in a updown circle sequence, the maximum speed doesn't exceed 2m/s. The ATEs of AI-IO and AirIO are 1.005 and 1.808 respectively, and AI-IO outperforms AirIO by 44.4\% in ATE.}
    \label{fig:low-spd-traj}
\end{figure}

As shown in TABLE \ref{our comparison}, IMU preintegration has huge cumulative errors in both velocity and position, and IMO is completely ineffective for flight data on non-fixed tracks. AirIO and AI-IO have good generalization properties, achieving high accuracy across diverse datasets using only a subset of random flight data. However, our AI-IO method significantly outperforms other methods, achieving 57.4\% and 29.3\% improvements on AVE and ATE, respectively, compared to the state-of-the-art method, AirIO. Fig. \ref{fig:high-spd-traj} and \ref{fig:low-spd-traj} show the velocity comparison and flight trajectory of a manual high and updown circle sequence, respectively. Whether comparing estimated velocity or position, AI-IO outperforms AirIO. The IMU preintegration and IMO divergence is too severe and is no longer shown.
\begin{table*}[ht]
\caption{Comparison on our datasets. The best results are bold.}
    \centering
    \setlength{\tabcolsep}{4.5pt}
    \renewcommand{\arraystretch}{1.2}
        \begin{tabular}{l|cccc|cccc|cccc|cccc}
        \toprule
        \multirow{2}{*}{\textbf{Sequence}} & 
        \multicolumn{4}{c|}{\textbf{IMU preintegration}$^1$} & 
        \multicolumn{4}{c|}{\textbf{IMO}$^1$} & 
        \multicolumn{4}{c|}{\textbf{AirIO Net}$^2$} & 
        \multicolumn{4}{c}{\textbf{AI-IO Net}$^2$}\\
        & AVE & RVE & ATE & RTE & AVE & RVE & ATE & RTE & AVE & RVE & ATE & RTE & AVE & RVE & ATE & RTE \\
        \midrule
        manual high     & 54.68 & 12.01 & 104.1 & 24.61 & 6.033 & 8.572 & 22.72 & 12.91 & 1.093 & 1.511 & 2.514 & 1.891 & \textbf{0.387} & \textbf{0.537} & \textbf{0.849} & \textbf{0.699} \\
        manual medium     & 54.79 & 15.39 & 104.3 & 31.47 & 6.178 & 7.824 & 36.99 & 17.65 & 0.836 & 1.087 & 1.773 & 1.235 & \textbf{0.285} & \textbf{0.363} & \textbf{0.994} & \textbf{0.569} \\
        manual low     & 38.40 & 25.35 & 72.95 & 50.17 & 3.882 & 5.439 & 12.37 & 8.397 & 0.715 & 0.909 & 2.072 & 1.161 & \textbf{0.268} & \textbf{0.336} &\textbf{0.783} & \textbf{0.434} \\
        eight     & 56.09 & 7.991 & 108.2 & 17.69 & 3.987 & 3.473 & 80.61 & 14.32 & 0.274 & 0.379 & 2.720 & 0.550 & \textbf{0.181} & \textbf{0.254} & \textbf{2.167} & \textbf{0.494} \\
        updown eight   & 55.41 & 7.546 & 107.0 & 16.26 & 2.739 & 2.643 & 65.06 & 9.517 & 0.356 & 0.438 & 1.678 & 0.776 & \textbf{0.182} & \textbf{0.236} & \textbf{1.256} & \textbf{0.407} \\
        circle   & 56.11 & 6.538 & 108.3 & 14.34 & 3.770 & 4.343 & 68.32 & 15.57 & 0.307 & 0.417 & 2.134 & 0.801 & \textbf{0.180} & \textbf{0.220} & \textbf{1.694} & \textbf{0.524} \\
        updown circle   & 56.10 & 6.568 & 108.7 & 14.46 & 3.648 & 2.723 & 110.4 & 16.10 & 0.418 & 0.525 & 4.326 & 1.147 & \textbf{0.204} & \textbf{0.236} & \textbf{3.380} & \textbf{0.643} \\
        random   & 53.37 & 15.78 & 103.2 & 31.45 & 3.937 & 4.022 & 94.44 & 14.62 & 0.448 & 0.588 & \textbf{3.255} & 0.941 & \textbf{0.206} & \textbf{0.235} & 3.347 & \textbf{0.594} \\
        \midrule
        \textbf{Average}     & 53.12 & 12.15 & 102.1 & 25.06 & 4.272 & 4.880 & 61.37 & 13.63 & 0.556 & 0.732 & 2.559 & 1.063 & \textbf{0.237} & \textbf{0.287} & \textbf{1.809} & \textbf{0.546} \\
        \bottomrule
        \multicolumn{17}{l}{1 leverage ground truth orientation to transform the IMU data to world frame for inference.} \\
        \multicolumn{17}{l}{2 leverage ground truth orientation to transform the net output to world frame for inference.} \\
        \end{tabular} 
        \label{our comparison}
\end{table*}


\subsection{Real World Validation}
\begin{figure}
    \centering
    \includegraphics[width=\linewidth]{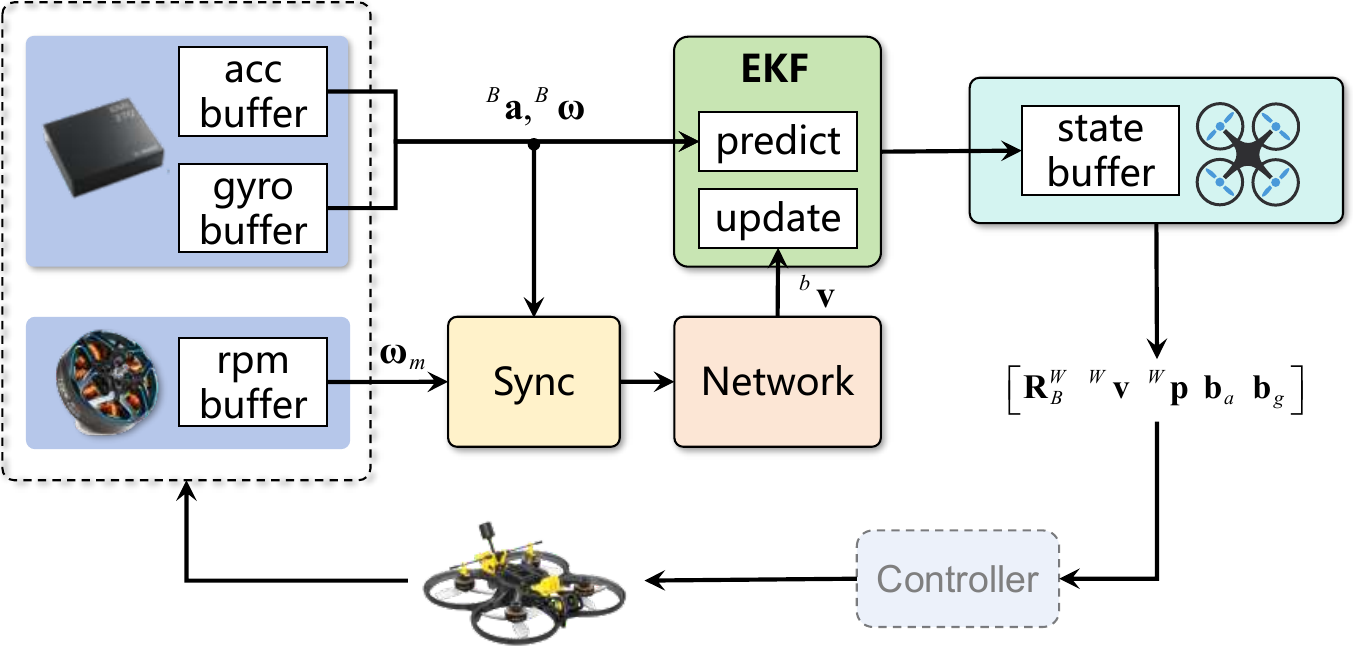}
    \caption{Real-time system overview: IMU-derived acceleration, angular velocity, and rotor speed (via bidirectional DShot) are time-synchronized and fed into the network for velocity prediction. Subsequently, EKF performs prediction and update steps, outputting pose information to the controller for closed-loop control.}
    \label{fig:system}
\end{figure}

To demonstrate how AI-IO supports real-world applications, we deploy it on the quadrotor platform shown in Fig. \ref{fig:dataset-platform}, which is equipped with a Radxa Zero3W computer (Quad-Core Arm Cortex-A55, up to 1.6GHz)\footnote{https://radxa.com/products/zeros/zero3w}. We design the real-time system demonstrated in Fig. \ref{fig:system} and then conduct indoor flight tests:

\subsubsection{How does AI-IO generalize to real-time systems?}
\begin{figure*}
    \centering
    \includegraphics[width=0.9\linewidth]{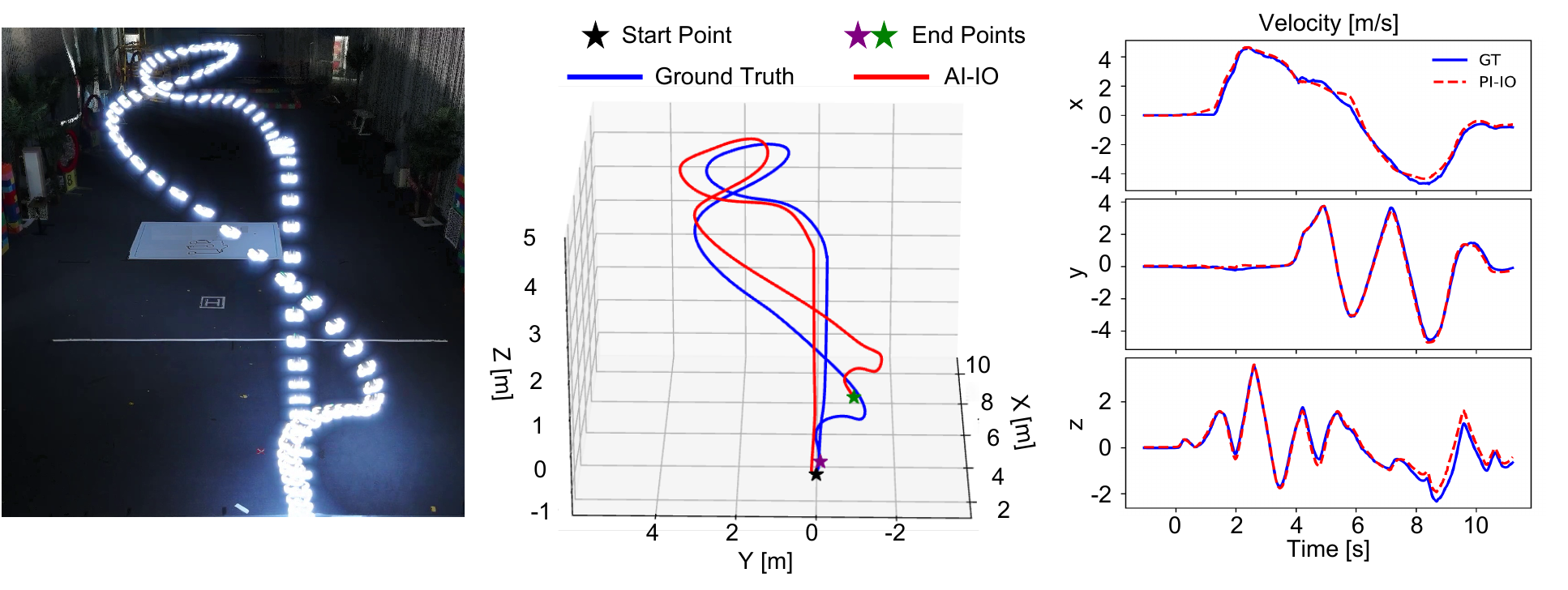}
    \caption{We deploy the trained lightweight network on a low-compute platform for real-time body-frame velocity estimation, then integrate it with an extended Kalman filter to estimate the pose. The quadrotor is maneuvered randomly by human pilots in low-light indoor environments.}
    \label{fig:realtime-overview}
\end{figure*}

The training data—randomly sampled by human pilots—covers trajectories with diverse characteristics, enabling AI-IO to naturally generalize to real-world flights. As shown in Fig.~\ref{fig:realtime-overview}, the deployed AI-IO achieves real-time pose estimation on a human-piloted quadrotor. The body-frame velocity prediction network runs with an average inference time of 8.9 ms while maintaining high accuracy. Pose estimation is performed via an EKF at 20 Hz, and the results remain highly consistent over long-range and agile trajectories.

\subsubsection{How physical inductive bias facilitate training?}
\begin{figure}
    \centering
    \includegraphics[width=\linewidth]{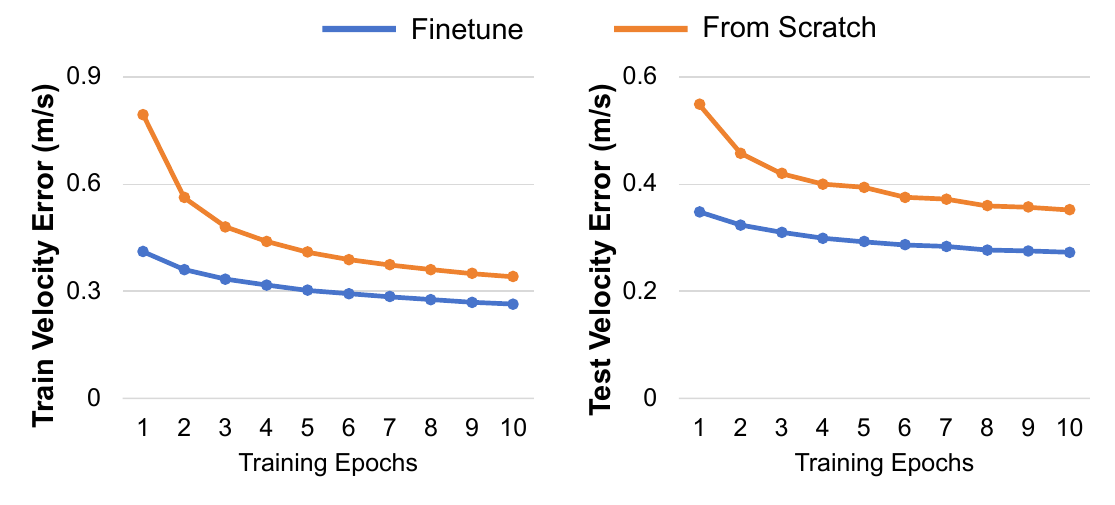}
    \caption{Comparison of velocity prediction errors between training from scratch and fine-tuning on the training and testing datasets.}
    \label{fig:finetune}
\end{figure}

To investigate the inductive bias of physical information learned during network training, we compare the velocity prediction errors of training from scratch versus fine-tuning. First, we train AI-IO on the dataset mentioned before. We then collect a small volume of 7-minute flight data using a quadrotor with modified mass and appearance, which is used to fine-tune the pre-trained model. As shown in Fig. \ref{fig:finetune}, compared to training from scratch with limited data, fine-tuning demonstrates higher data utilization efficiency and converges to lower errors more rapidly by leveraging the physical information embedded in the pre-trained base model. This indicates that for any quadrotor, only a small amount of data is required to quickly train a usable inertial odometry.

\subsubsection{How does AI-IO perform compared to other methods?}
We mount the T265 camera on the original quadrotor platform and evaluate AI-IO (fine-tuned as described previously) by comparing its localization accuracy with state-of-the-art AirIO (fine-tuned in the same way) and the T265’s VIO in indoor flight scenarios. To mitigate potential degradation in the T265’s VIO, we place textured objects on one side of the environment and then turn off the lights during flight. As shown in Fig. \ref{fig:io-vio}, AI-IO consistently outperforms AirIO. Before the lights are turned off, AI-IO achieves localization accuracy comparable to the T265’s VIO. After light extinction, however, the T265’s VIO degrades and diverges from the ground truth, whereas AI-IO maintains high consistency.




\section{CONCLUSION \& DISCUSSION}
In this work, we propose a novel perspective grounded in quadrotor aerodynamics and IMU models to explain and guide the design of learning-based inertial odometry. Guided by this framework, we develop the AI-IO, a lightweight transformer-based inertial odometry that, when combined with an EKF, achieves real-time inference on real-world and delivers high pose estimation accuracy. Additionally, we collect and release a high-maneuverability IMU dataset for quadrotors. Experimental results across diverse datasets and real-world platforms demonstrate that our framework achieves a 49.4\% improvement in velocity estimation and a 30.9\% improvement in position estimation compared to state-of-the-art methods. Future work may focus on developing network architectures more tightly integrated with physical priors (e.g., PINNs) to enhance interpretability and enable integration with multi-sensor frameworks such as VIO for further improvement in position estimation.



\bibliographystyle{ieeetr}
\bibliography{reference}

\end{document}